\documentclass{article}
\usepackage{spconf,amsmath,graphicx}
\usepackage{times}
\usepackage{graphicx}
\usepackage{subfigure}
\usepackage{color}
\usepackage{indentfirst}
\usepackage[linesnumbered,boxed]{algorithm2e}
\usepackage{algorithmic}
\usepackage{booktabs}
\usepackage{stfloats}
\usepackage{amsmath}

\title{Cascaded Self-Representation for Outlier Detection}
\name{Qi Yang, Hao Zhu}
\address{Vector Lab, JD Finance, Beijing}
\begin{document}
%
\maketitle
%


\begin{abstract}
Many methods based on sparse and low-rank representation been developed along with guarantees of correct outlier detection.
Self-representation states that a point in a subspace can always be expressed as a linear combination of other points in the subspace. 
A suitable Markov Chain can be defined on the self-representation and it allows us to recognize the difference between inliers and outliers.
However, the reconstruction error of self-representation that is still informative to detect outlier detection, is neglected.
Inspired by the gradient boosting, in this paper, we propose a new outlier detection framework that combines a series of weak "outlier detectors" into a single strong one in an iterative fashion by constructing multi-pass self-representation. 
At each stage, we construct a self-representation based on elastic-net and define a suitable Markov Chain on it to detect outliers.
The residual of the self-representation is used for the next stage to learn the next weaker outlier detector. 
Such a stage will repeat many times. And the final decision of outliers is generated by the previous all results.
Experimental results on image and speaker datasets demonstrate its superiority with respect to state-of-the-art sparse and low-rank outlier detection methods. 


\end{abstract}

\begin{keywords}
Outlier Detection, Self-Representation, Sparse Coding
\end{keywords}
\section{Introduction}
Outlier detection, also called anomaly detection, identifies unusual data patterns that are different from the majority of data.
These unexpected patterns are always called anomalies or outliers, the ability to detect anomalies has significant and critical help, and anomalies often provides useful information in various application domains, such as intrusion detection, fraud detection, fault detection, suspicious transaction detection and abnormal moving activity detection, etc. Anomaly detection can be roughly categorized into three ways: statistical, algebraic and self-representation based.

RANdom SAmple Consensus (RANSAC) \cite{fischler1981random}, as a statistical method for outlier detection, employs a sampling strategy to find a subspace that fits as many data as possible iteratively. In this processing, data points are removed from the dataset to find a new subspace until a given threshold on the percents of inliers is reached.
Although theoretically, it is able to find correct subspace in noisy data even in the presence of outliers, the computational cost is still a challenge. Its variants like \cite{torr1998robust} and \cite{torr2002bayesian} still face the similar problem.

Algebraic methods have been to robustly learn the subspaces by penalizing the sum of an unsquared distance of points to the closest subspace \cite{lerman2011robust,zhang2012hybrid} compared with Principal Component Analysis (PCA) minimizing the sum of square error. Thus they are robust to outliers because of reducing the contributions from large residuals arising from outliers. However, the optimization problem is usually nonconvex and a good initialization is extremely important for finding the optimal solution. 
Recently the PCA problem with robustness have been solved to corrupted entries \cite{wright2009robust}, which has led to many recent methods for PCA with robustness to outliers. 
A prominent advantage of convex optimization techniques is that they are guaranteed to correctly identify outliers under certain conditions. 
Nonetheless, these methods typically model a unique inlier subspace, e.g., by a low-rank constraint in Outlier Pursuit \cite{xu2010robust}, and therefore cannot deal with multiple inlier subspaces since the union of multiple subspaces could be high-dimensional.

An alternative to algebraic approaches are methods that exploit the self-expressiveness of subspaces. In most of these approaches, the goal becomes finding a clustering that renders a suitably defined affinity matrix, typically obtained by solving a sparsification \cite{elhamifar2009sparse} or rank minimization problem \cite{liu2010robust}. These methods work well in the presence of noise and offer theoretical recovery guarantees. On the other hand, handling outliers requires augmenting the objective function with additional regularization terms, whose parameters must be usually hand-tuned to obtain good performance and typically recovery guarantees are lost. Based on self-representation, \cite{you2017provable} design a random walk process and identify outliers as those whose probabilities tend to zero, which is parameter-free. However, the residual of self-representation which is neglected is still informative and contains clues to make outlier detection better.


In this paper, we introduce a novel cascade architecture that aims to extend the random walk based outlier detection onto hierarchical self-representations in a principled way. 
To achieve the information from the reconstruction error in a linear combination, we present a multi-stages cascade framework for self-representation based on sparse coding with Elastic Net \cite{zou2005regularization}.
The residuals at a layer is computed by the difference between the original input and the aggregated reconstructions of the previous layers.
And then the residuals at a layer can be represented with self-representation just like the original inputs did.
Multi self-representation can be used to detect outliers with a random work based method, the final decision of outliers are fused by a linear combination of different results from different layers.
%





\section{Related work}
Motivated by the observation that outliers do not have sparse representations, \cite{soltanolkotabi2012geometric} declare a point as an outlier if its $\ell1$ norm of representation is above a threshold. However, this $\ell1$-thresholding strategy is not robust to outliers that is not good enough to discriminate outliers since their representation vectors may have many small small $\ell1$-norms. Low-Rank Representation (LRR) \cite{liu2010robust} employs sparsity assumption (i.e. $\ell_{2,1}$-norm) on the reconstruction error of low rank self-representation to learn a robust model, and the column-sparse matrix rather than the low-rank matrix can be used to indicate the outliers.



In \cite{moonesinghe2008outrank}, a graph-based outlier detection framework is proposed. The main idea is to represent the underlying dataset as a weighted undirected graph and then use the random work on the graph to find outliers. The random walk model is designed to find nodes that are most “central” to the graph and thus the outliers are far from the "central" can be efficiently detected. \cite{you2017provable} extends graph construction from kernel based similarity to sparse representation based self-representation that inliers express themselves with only other inliers when they lie in a union of low dimensional subspaces. To avoid the non-convergence because of directed graph, researcher choose to calculate the average of the first $T$ t-step probability distributions.


\begin{figure}[!htbp]
\centering
\includegraphics[width=1\linewidth]{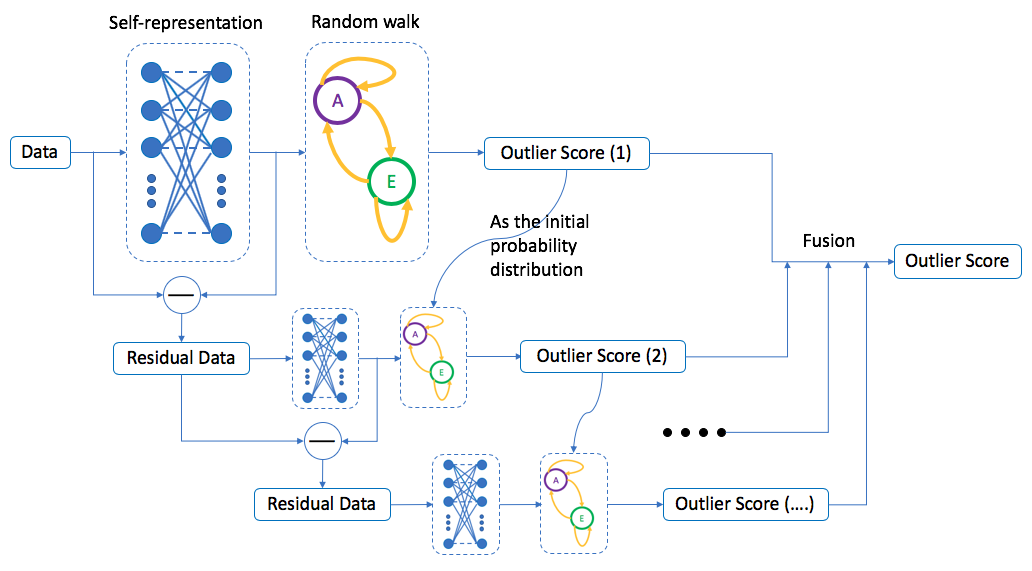}
\caption{The Framework of Cascaded Self-Representation for Outlier Detection}
\label{fig:Ensemble}
\end{figure}

\section{Method}
In this section, we present cascaded self-representation based outlier detection method shown in Fig.~\ref{fig:Ensemble}. We first describe the data self-representation and a random walk algorithm on the graph based on the representation to identify the sets of inliers and outliers. Then, we introduce a cascaded architecture that calculates self-representation from the resulting residuals of the previous layers. Finally we ensemble the all the results into one set of scores to identify the inliers and outliers.

\subsection{Elastic Net based Self-Representation}

Given unlabeled data points $\{X\}_{i=1,...,N}$ drawn from multiple linear subspaces $\{S\}_{i=1,...,K}$, one can express a point in a subspace as a linear combination of other points in the same subspace, and the data contains inliers and outliers. 
We can get the self-representation coefficient matrix by the optimization problem:
\begin{equation}
\quad\min\limits_{C}\frac{\gamma}{2}\|X-XC\|^2_2 + \lambda\|C\|_{1} + \frac{1-\lambda}{2}\|C\|_2^2 \quad s.t.\; C_{ii} = 0 \\
\label{equ:elastic}
\end{equation}
where $C$ is the self-representation coefficient matrix, $\gamma > 0$ and $\lambda\in[0,1)$ , the optional diagonal constraint on $C$ prevents trivial solutions for sparsity-inducing norms, such as the $\ell1$ norm. Compared with the $\ell1$ norm constraint, A mixture of $\ell1$ and $\ell2$ norms is able to balance the subspace preserving and connectedness properties. However, the residual term $\|X-XC\|_2^2$, also called reconstruction error, is ignored by the later random walk based outlier detection.

\subsection{Graph-based Outlier Detection}

With a Self-Representation $C$, we can get the directed weighted graph $G$: The vertices of $G$ correspond to the data points $X$, and the edges are given by the (weighted) adjacency matrix $A = |C|^T$.
In this graph $G$, inliers only have connections with the inliers while the outliers have edges with both inliers and outliers, which we can use random walk to detect the outliers. 
The transition probability from $x_i$ to $x_j$ at the next time is given by $p_{ij} = a_{ij} /d_i$ with $d_i = \sum_{j} a_{ij}$, where $p_{ij}\in P$ and $a_{ij}\in A$ . By this definition, if the starting point of a random walk is an inlier then it will never escape the set of inliers as there is no edge going from any inlier to any outlier. In contrast, a random walk starting from an outlier will likely end up in an inlier state since once it enters any inlier it will never return to an outlier state. Thus, by using different data points to initialize random walks, outliers can be identified by observing the final probability distribution of the state of the random walks.
A uniform distribution $\pi^{(0)}$ = [$\frac{1}{N}$,...,$\frac{1}{N}$] is defined as the initial probability distribution, we define the formula:
\begin{equation}
    \hat{\pi}^{(T)} = \frac{1}{T}\sum_{t=1}^{T}\pi^{(0)} P^t
    \label{equ:RW}
\end{equation}
where the average of the t-step can ensure the convergence . It is expected that eventually the inliers have high probabilities states and outliers have low probabilities, so we can use the threshold value $\epsilon$ to identify the ouliers if $\hat{\pi}^{(T)} \leq \epsilon$.

\subsection{Outlier Detection by Cascaded Self-Representation}

As mentioned above, the method of self-representation based outlier detection often formulates the problem at hand using a linear model with regularization by $\ell1$ and $\ell2$-norms, but the reconstruction error is ignored that hinders exploiting self-representation in their full potential.
In comparison, our approach exploits a recursive way where we encode self-representation with the resulting residuals of the stage in previous stages.
In a single stage, we represent the residual as a linear combination of other points, where we keep the same as single stage.
Let $\hat{X}^n$ denote the estimated $n$-th stage and $X$ denote unlabeled data points, then the overall process can be described as:
\begin{equation}
    X = \hat{X}^0 + \hat{X}^1 + ...\hat{X}^n\quad s.t.\; \hat{X}^i = (X-\sum_{j=0}^{i-1}\hat{X}^j)C^i,\; C^i_{jj}=0
\label{equ:cascaded}
\end{equation}
where $\hat{X}^0$ is a matrix with zero entries which is used to easily formulate our motivation with the Equ.~\ref{equ:cascaded}.

The flow diagram of our cascade framework is showed as Fig.1 and the algorithm is Alg.~\ref{alg:1}. Given the data $X$, we first get the self-representation $C^1$  and the transition probability, then run Equ.~\ref{equ:RW} on it to get the outlier score $S^1$, for the $\hat{X}^1$, it can be solve by $\hat{X}^1 = XC^1$. And then we can use the $X-\hat{X}^1$ do again the self-representation and random-walk to get the outlier score $S^2$, we know that the outlier score $S^1$ get from the random-walk's initial probability distribution is $\pi^{(0)}$, however, the outlier score $S^2$ solved by the same process's initial probability distribution is $S^1$. The rest can be done in the same manner to get the $\hat{C}^2$, $\hat{C}^3$,...,$\hat{C}^n$, and the residuals $X-\sum_{j=0}^{i-1}\hat{X}^j$, and the $S^3$, $S^4$,...,$S^n$. After all is done, do the score fusion of the  $S^1$,...,$S^n$ to get the final Outlier Score $S$, and identify the Outliers if $S \leq \epsilon$.

\begin{algorithm}[htb]  
	\renewcommand{\algorithmicrequire}{\textbf{Input:}}
	\renewcommand{\algorithmicensure}{\textbf{Output:}}
	\caption{Outlier Detection by Cascaded Self-Representation}
	\label{alg:1}
 	\begin{algorithmic}[1]
		\REQUIRE Unlabeled data points $X$, the threshold value $\epsilon$
		\ENSURE Outlier score $S$ of each points and Outliers by $\epsilon$
		\STATE Initial probability distribution $\pi^{(0)} = [\frac{1}{N},...,\frac{1}{N}]$
		\FOR{$i = 1$ to $n$}
		\IF{$i = 1$}
		\STATE $C^1$ $\leftarrow$ solved by Equ.~\ref{equ:elastic} 
		\STATE Outlier Score $S^1$  $\leftarrow$ solved by Equ.~\ref{equ:RW} with $\pi^{(0)}$
		\STATE $\hat{X}^1$ $\leftarrow$ solved by Equ.~\ref{equ:cascaded}
		\ENDIF
		\STATE Update probability distribution  = Outlier Score $S^{(i-1)}$
		\STATE $C^i$ $\leftarrow$ solved by Equ.~\ref{equ:elastic} with residuals $X-\sum_{j=0}^{i-1}\hat{X}^j$
		\STATE Outlier Score $S^i$  $\leftarrow$ solved by Equ.~\ref{equ:RW} with $S^{(i-1)}$
		\STATE $\hat{X}^i$ $\leftarrow$ solved by Equ.~\ref{equ:cascaded}
		\ENDFOR
		\STATE Score fusion result $S$ $\leftarrow$ from  $S^1,S^2,..,S^n$
		\STATE Identify the Outliers if $S \leq \epsilon$
		\RETURN{} Outlier score $S$ and Outliers 
 	\end{algorithmic}  
\end{algorithm}

\section{Experiments}
In this section, we use three different image datasets and one speaker dataset to evaluate the proposed method for outlier detection. In particular, we analyzed the performance results in details and compared with state-of-the-art techniques.
\subsection{Experimental setup}
We construct our method in matlab, and evaluated it on four publicly available databases, i.e., the Extended Yale B \cite{Georghiades:2001:FMI:378040.378083} , the Caltech-256 \cite{griffin2007caltech}, the Coil-100 object image dataset \cite{nene1996columbia} and the TIMIT Small dataset \cite{hibraj2018speaker}. For our method we set the number of iterations T to be 1000, and the number of stages is 3 since larger stages do not contribute better performance significantly.

We compare with 3 other representative methods that are designed for detecting outliers in one or multiple subspaces:  LRR \cite{liu2010robust} , $\ell$-thresholding \cite{soltanolkotabi2012geometric} and R-graph \cite{you2017provable}. All other methods are implemented according to the description in their respective papers. We call our method Outlier Detection by Cascaded Self-Representation (ODCSR) in this section.

For each outlier detection method, we use two metric to evaluate its performance. One is the area under the curve (AUC) as a metric of performance in terms of the ROC, the other is the F1-score, which is the harmonic mean of precision and recall. Notice that a numerical value for each data point that indicates its “outlierness” and a threshold value for determining inliers and outliers are required. 

\subsection{Extended Yale B Dataset}
The Extended Yale B dataset is a popular benchmark for subspace clustering. It contains frontal face images of 38 individuals each under 64 different illumination conditions. Following the experimental setup of \cite{elhamifar2009sparse}, we down-sampled the original face images from $192 \times 168$ to $42 \times 42$ pixels, which makes it computationally feasible for the baselines.

We designed a series of experiments to test the ability of the methods to deal with multiple inlier groups. For example, we randomly choose 1 or 3  individual from the 38 subjects and use all 64 images of the subjects as the inliers. Then the images from the remaining 37 or 35 subjects as outliers with at most one image from each subject. The overall data set has 35\% or 15\%  outliers. The results of this experiment are reported in Table~\ref{tab:yaleB}. The last column have the results of our methods, note that in each row the best result is typeset in bold. We can see that our cascade subspace clustering method is the best method. 


\linespread{0.5}
\begin{table}[htb]
  \centering
  \caption{Results on the Extended Yale B dataset} 
    \begin{tabular}{cccccccccc}
    \\
    \toprule
          & \multicolumn{1}{l}{LRR} &  \multicolumn{1}{l}{$\ell1$-thresholding} & \multicolumn{1}{l}{R-graph} & \multicolumn{1}{l}{\textbf{ODCSR(ours)}} \\
    \midrule
    \multicolumn{5}{l}{\textit{Inliers:from \textbf{one} subject, Outliers:35\% from other subjects}} \\
    \midrule
    AUC   & 0.857  & 0.844 & 0.986 & \textbf{0.990} \\
    F1    & 0.797  & 0.763 & 0.951 & \textbf{0.956} \\
    \midrule
    \multicolumn{5}{l}{\textit{Inliers:from \textbf{three} subjects, Outliers:15\% from other subjects}} \\
    \midrule
    AUC   & 0.807 & 0.848 & 0.985 & \textbf{0.986} \\
    F1    & 0.509 & 0.545 & 0.878 & \textbf{0.886} \\
    \bottomrule
    \end{tabular}%
  \label{tab:yaleB}%
\end{table}%

\subsection{Caltech-256 Dataset}

For Caltech-256, which contains 256 object categories with a total of 30,607 images, and each category has at least 80 images. In our experiments, images in Caltech-256 are represented by a 4,096-dimensional feature vector extracted from the last fully connected layer of the 16-layer VGG network \cite{simonyan2014very}. Like the above, images in $n \in \{1, 3, 5\}$ randomly chosen categories and the first 150 images in it are used as inliers, and we randomly pick a certain number of images from the “clutter” category as outliers such that there are 50\% outliers in each experiment.

The results of this experiment are reported in Table~\ref{tab:Caltech}. The last column have the results of our methods, note that in each row the best result is typeset in bold, and we are the best method. 

\linespread{0.5}
\begin{table}[htb]
  \centering
  \caption{Results on the Caltech-256 dataset} 
    \begin{tabular}{cccccccccc}
    \\
    \toprule
          & \multicolumn{1}{l}{LRR}  & \multicolumn{1}{l}{$\ell1$-thresholding} & \multicolumn{1}{l}{R-graph} & \multicolumn{1}{l}{\textbf{ODCSR(ours)}} \\
    \midrule
    \multicolumn{5}{l}{\textit{Inliers:from \textbf{one} category, Outliers:50\% from 257-clutter}} \\
    \midrule
    AUC   & 0.907 & 0.772 & 0.948 & \textbf{0.983} \\
    F1    & 0.893 & 0.772 & 0.914 & \textbf{0.946} \\
    \midrule
    \multicolumn{5}{l}{\textit{Inliers:from \textbf{three} categories, Outliers:50\% from 257-clutter}} \\
    \midrule
    AUC   & 0.479 & 0.810 & 0.929 & \textbf{0.984} \\
    F1    & 0.671 & 0.782 & 0.880 & \textbf{0.947} \\
    \midrule
    \multicolumn{5}{l}{\textit{Inliers:from \textbf{five} categories, Outliers:50\% from 257-clutter}} \\
    \midrule
    AUC   & 0.337 & 0.774 & 0.913 & \textbf{0.984} \\
    F1    & 0.667 & 0.762 & 0.858 & \textbf{0.952} \\
    \bottomrule
    \end{tabular}%
  \label{tab:Caltech}%
\end{table}%

\subsection{Coil-100 Dataset}

The Coil-100 dataset contains 7,200 images of 100 different objects. Each object has 72 images taken at pose intervals of 5 degrees, with the images being of size $32 \times 32$. For Coil-100, we randomly pick $n \in \{1, 4, 7\}$ categories as inliers and pick at most one image from each of the remaining categories as outliers. The results of this experiment are reported in Table~\ref{tab:Coil}. And we have the best performance.

\linespread{0.5}
\begin{table}[!htbp]
\centering
  \caption{Results on the Coil-100  dataset}
    \begin{tabular}{cccccccccc}
    \\
    \toprule
          & \multicolumn{1}{l}{LRR} & \multicolumn{1}{l}{$\ell1$-thresholding} & \multicolumn{1}{l}{R-graph} & \multicolumn{1}{l}{\textbf{ODCSR(ours)}} \\
    \midrule
    \multicolumn{5}{l}{\textit{Inliers:from \textbf{one} subject, Outliers:50\% from other subjects}} \\
    \midrule
    AUC   & 0.847 & 0.991 & 0.997 & \textbf{0.999} \\
    F1    & 0.872 & 0.978 & 0.900 & \textbf{0.995} \\
    \midrule
    \multicolumn{5}{l}{\textit{Inliers:from \textbf{four} subjects, Outliers:25\% from other subjects}} \\
    \midrule
    AUC   & 0.687 & 0.992 & 0.996 & \textbf{0.998} \\
    F1    & 0.541 & 0.941 & 0.970 & \textbf{0.981} \\
    \midrule
    \multicolumn{5}{l}{\textit{Inliers:from \textbf{seven} subjects, Outliers:15\% from other subjects}} \\
    \midrule
    AUC   & 0.628 & 0.991 & 0.996 & \textbf{0.997} \\
    F1    & 0.366 & 0.897 & 0.955 & \textbf{0.963} \\
    \bottomrule
    \end{tabular}%
  \label{tab:Coil}%
\end{table}%

\subsection{TIMIT Small Dataset}

The TIMIT dataset is composed of 6,300 phrases (10 phrases per speaker), spoken by 438 males (70\%) and 192 females (30\%). In our experiment, we used the same 40 speakers dataset as reported by these earlier attempts (here called TIMIT Small). we randomly pick $n \in \{3, 5, 7\}$ speakers as inliers and pick at most one phrase from each of the remaining speakers as outliers. The results of this experiment are reported in Table~\ref{tab:TIMIT}. And we can see that our method is still the best.


\linespread{0.5}
\begin{table}[!htbp]
\centering
  \caption{Results on the TIMIT Small dataset}
    \begin{tabular}{cccccccccc}
    \\
    \toprule
          & \multicolumn{1}{l}{LRR} & \multicolumn{1}{l}{$\ell1$-thresholding} & \multicolumn{1}{l}{R-graph} & \multicolumn{1}{l}{\textbf{ODCSR(ours)}} \\
    \midrule
    \multicolumn{5}{l}{\textit{Inliers:from \textbf{three} speakers, Outliers:15\% from other speakers}} \\
    \midrule
    AUC   & 0.867 &0.873  & 0.959 & \textbf{0.972}\\
    F1    & 0.698 & 0.660 & 0.837 & \textbf{0.867} \\
    \midrule
    \multicolumn{5}{l}{\textit{Inliers:from \textbf{five} speakers, Outliers:15\% from other speakers}} \\
    \midrule
    AUC   & 0.847 & 0.839 & 0.944 & \textbf{0.951} \\
    F1    & 0.630 & 0.598 & 0.774 & \textbf{0.814} \\
    \midrule
    \multicolumn{5}{l}{\textit{Inliers:from \textbf{seven} speakers, Outliers:15\% from other speakers}} \\
    \midrule
    AUC   & 0.813 & 0.826 & 0.936 & \textbf{0.944} \\
    F1    & 0.536 & 0.557 & 0.749 & \textbf{0.759} \\
    \bottomrule
    \end{tabular}%
  \label{tab:TIMIT}%
\end{table}%

\section{Conclusion}


In this paper, we have proposed a general framework for outlier detection in a manner of cascade subspace clustering. 
Our architecture consists of multi-stages, each one is random walk outlier detection based on self-representation with the residual of last layers. 
Eventually, each result from different paths is fused to make the final decision of outliers.
%
%
Our experiments have demonstrated that our cascade method provides significant improvement over the state-of-the-art outlier detection solutions in terms of AUC and F1 on several datasets.

\bibliographystyle{IEEEbib}
\bibliography{egbib}
\noindent 

\end{document}